# Speeding Up the Convergence of Value Iteration in Partially Observable Markov Decision Processes

**Nevin L. Zhang**                                LZHANG@CS.UST.HK
**Weihong Zhang**                                 WZHANG@CS.UST.HK
*Department of Computer Science*
*Hong Kong University of Science & Technology*
*Clear Water Bay Road, Kowloon, Hong Kong, CHINA*

## Abstract

Partially observable Markov decision processes (POMDPs) have recently become popular among many AI researchers because they serve as a natural model for planning under uncertainty. Value iteration is a well-known algorithm for finding optimal policies for POMDPs. It typically takes a large number of iterations to converge. This paper proposes a method for accelerating the convergence of value iteration. The method has been evaluated on an array of benchmark problems and was found to be very effective: It enabled value iteration to converge after only a few iterations on all the test problems.

## 1. Introduction

POMDPs model sequential decision making problems where effects of actions are nondeterministic and the state of the world is not known with certainty. They have attracted many researchers in Operations Research and Artificial Intelligence because of their potential applications in a wide range of areas (Monahan 1982, Cassandra 1998b), one of which is planning under uncertainty. Unfortunately, there is still a significant gap between this potential and actual applications, primarily due to the lack of effective solution methods. For this reason, much recent effort has been devoted to finding efficient algorithms for POMDPs (e.g., Parr and Russell 1995, Hauskrecht 1997b, Cassandra 1998a, Hansen 1998, Kaelbling *et al.* 1998, Zhang *et al.* 1999).

Value iteration is a well-known algorithm for POMDPs (Smallwood and Sondik 1973, Puterman 1990). It starts with an initial value function and iteratively performs dynamic programming (DP) updates to generate a sequence of value functions. The sequence converges to the optimal value function. Value iteration terminates when a predetermined convergence condition is met.

Value iteration performs typically a large number of DP updates before it converges and DP updates are notoriously expensive. In this paper, we develop a technique for reducing the number of DP updates.

DP update takes (the finite representation of) a value function as input and returns (the finite representation of) another value function. The output value function is closer to the optimal than the input value function. In this sense, we say that DP update *improves* its input. We propose an approximation to DP update called *point-based DP update*. Point-based DP update also improves its input, but possibly to a lesser degree than standard DP update. On the other hand, it is computationally much cheaper. During value iteration, we perform point-based DP update a number of times in between two standard DP updates. The number of standard DP updates can be reduced this way since point-based DP update improves its input. The reduction does not come with a high cost since point-based DP update takes little time.

The rest of this paper is organized as follows. In the next section we shall give a brief review of POMDPs and value iteration. The basic idea behind point-based DP update will be explained in Section 3. After some theoretical preparations in Section 4, we shall work out the details of point-based DP update in Section 5. Empirical results will be reported in Section 6 and possible variations evaluated in Section 7. Finally, we shall discuss related work in Section 8 and provide some concluding remarks in Section 9.

## 2. POMDPs and Value Iteration

### 2.1 POMDPs

A *partially observable Markov decision process* (POMDP) is a sequential decision model for an agent who acts in a stochastic environment with only partial knowledge about the state of its environment. The set of possible states of the environment is referred to as the *state space* and is denoted by $\mathcal{S}$. At each point in time, the environment is in one of the possible states. The agent does not directly observe the state. Rather, it receives an observation about it. We denote the set of all possible observations by $\mathcal{Z}$. After receiving the observation, the agent chooses an action from a set $\mathcal{A}$ of possible actions and executes that action. Thereafter, the agent receives an immediate reward and the environment evolves stochastically into a next state.

Mathematically, a POMDP is specified by: the three sets $\mathcal{S}$, $\mathcal{Z}$, and $\mathcal{A}$; a *reward function* $r(s, a)$; a *transition probability function* $P(s'|s, a)$; and an *observation probability function* $P(z|s', a)$. The reward function characterizes the dependency of the immediate reward on the current state $s$ and the current action $a$. The transition probability characterizes the dependency of the next state $s'$ on the current state $s$ and the current action $a$. The observation probability characterizes the dependency of the observation $z$ at the next time point on the next state $s'$ and the current action $a$.

### 2.2 Policies and Value Functions

Since the current observation does not fully reveal the identity of the current state, the agent needs to consider all previous observations and actions when choosing an action. Information about the current state contained in the current observation, previous observations, and previous actions can be summarized by a probability distribution over the state space (Aström 1965). The probability distribution is sometimes called a *belief state* and denoted by $b$. For any possible state $s$, $b(s)$ is the probability that the current state is $s$. The set of all possible belief states is called the *belief space*. We denote it by $\mathcal{B}$.

A *policy* prescribes an action for each possible belief state. In other words, it is a mapping from $\mathcal{B}$ to $\mathcal{A}$. Associated with a policy $\pi$ is its *value function* $V^{\pi}$. For each belief state $b$, $V^{\pi}(b)$ is the expected total discounted reward that the agent receives by following the policy starting from $b$, that is

$$V^\pi(b) = E_{\pi, b}[\sum_{t=0}^{\infty} \lambda^t r_t], \tag{1}$$

where $r_t$ is the reward received at time $t$ and $\lambda$ ($0 \leq \lambda < 1$) is the *discount factor*. It is known that there exists a policy $\pi^*$ such that $V^{\pi^*}(b) \geq V^\pi(b)$ for any other policy $\pi$ and any belief state $b$ (Puterman 1990). Such a policy is called an *optimal policy*. The value function of an optimal policy is called the *optimal value function*. We denote it by $V^*$. For any positive number $\epsilon$, a policy $\pi$ is *$\epsilon$-optimal* if

$$V^\pi(b) + \epsilon \geq V^*(b) \quad \forall b \in \mathcal{B}.$$

## 2.3 Value Iteration

To explain value iteration, we need to consider how belief state evolves over time. Let $b$ be the current belief state. The belief state at the next point in time is determined by the current belief state, the current action $a$, the next observation $z$. We denote it by $b_z^a$. For any state $s'$, $b_z^a(s')$ is given by

$$b_z^a(s') = \frac{\sum_s P(z, s'|s, a) b(s)}{P(z|b, a)}, \tag{2}$$

where $P(z, s'|s, a) = P(z|s', a) P(s'|s, a)$ and $P(z|b, a) = \sum_{s, s'} P(z, s'|s, a) b(s)$ is the renormalization constant. As the notation suggests, the constant can also be interpreted as the probability of observing $z$ after taking action $a$ in belief state $b$.

Define an operator $T$ that takes a value function $V$ and returns another value function $TV$ as follows:

$$TV(b) = \max_a [r(b, a) + \lambda \sum_z P(z|b, a) V(b_z^a)] \quad \forall b \in \mathcal{B} \tag{3}$$

where $r(b, a) = \sum_s r(s, a) b(s)$ is the expected immediate reward for taking action $a$ in belief state $b$. For a given value function $V$, a policy $\pi$ is said to be *$V$-improving* if

$$\pi(b) = \arg \; \max_a [r(b, a) + \lambda \sum_z P(z|b, a) V(b_z^a)] \quad \forall b \in \mathcal{B}. \tag{4}$$

Value iteration is an algorithm for finding $\epsilon$-optimal policies. It starts with an initial value function $V_0$ and iterates using the following formula:

$$V_n = TV_{n-1}.$$

It is known (e.g., Puterman 1990, Theorem 6.9) that $V_n$ converges to $V^*$ as $n$ goes to infinity. Value iteration terminates when the *Bellman residual* $\max_b |V_n(b) - V_{n-1}(b)|$ falls below $\epsilon(1 - \lambda)/2\lambda$. When it does, a $V_n$-improving policy is $\epsilon$-optimal (e.g., Puterman 1990).

Since there are infinitely many belief states, value functions cannot be explicitly represented. Fortunately, the value functions that one encounters in the process of value iteration admit implicit finite representations. Before explaining why, we first introduce several technical concepts and notations.

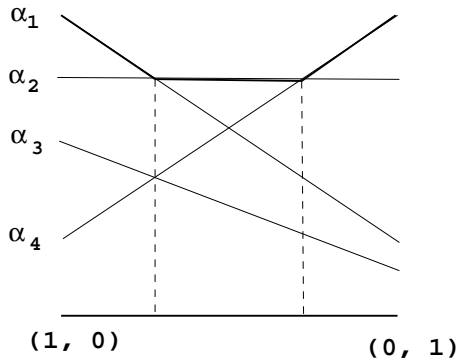

Figure 1: Illustration of Technical Concepts.

## 2.4 Technical and Notational Considerations

For convenience, we view functions over the state space vectors of size $|\mathcal{S}|$. We use lower case Greek letters $\alpha$ and $\beta$ to refer to vectors and script letters $\mathcal{V}$ and $\mathcal{U}$ to refer to sets of vectors. In contrast, the upper case letters $V$ and $U$ always refer to value functions, that is functions over the belief space $\mathcal{B}$. Note that a belief state is a function over the state space and hence can be viewed as a vector.

A set $\mathcal{V}$ of vectors *induces* a value function as follows:

$$f(b) = \max_{\alpha \in \mathcal{V}} \alpha \cdot b \quad \forall b \in \mathcal{B},$$

where $\alpha \cdot b$ is the inner product of $\alpha$ and $b$, that is $\alpha \cdot b = \sum_s \alpha(s) b(s)$. For convenience, we shall abuse notation and use $\mathcal{V}$ to denote both a set of vectors and the value function induced by the set. Under this convention, the quantity $f(b)$ can be written as $\mathcal{V}(b)$.

A vector in a set is *extraneous* if its removal does not affect the function that the set induces. It is *useful* otherwise. A set of vectors is *parsimonious* if it contains no extraneous vectors.

Given a set $\mathcal{V}$ and a vector $\alpha$ in $\mathcal{V}$, define the *open witness region* $R(\alpha, \mathcal{V})$ and *closed witness region* $\overline{R}(\alpha, \mathcal{V})$ of $\alpha$ w.r.t $\mathcal{V}$ to be regions of the belief space $\mathcal{B}$ respectively given by

$$R(\alpha, \mathcal{V}) = \{b \in \mathcal{B} | \alpha \cdot b > \alpha' \cdot b, \ \forall \alpha' \in \mathcal{V} \backslash \{\alpha\}\}$$
$$\overline{R}(\alpha, \mathcal{V}) = \{b \in \mathcal{B} | \alpha \cdot b \geq \alpha' \cdot b, \ \forall \alpha' \in \mathcal{V} \backslash \{\alpha\}\}$$

In the literature, a belief state in the open witness region $R(\alpha, \mathcal{V})$ is usually called a *witness point* for $\alpha$ since it testifies to the fact that $\alpha$ is useful. In this paper, we shall call a belief state in the closed witness region $\overline{R}(\alpha, \mathcal{V})$ a *witness point* for $\alpha$.

Figure 1 diagrammatically illustrates the aforementioned concepts. The line at the bottom depicts the belief space of a POMDP with two states. The point at the left end represents the probability distribution that concentrates all its masses on one of the states, while the point at the right end represents the one that concentrates all its masses on the other state. There are four vectors $\alpha_1$, $\alpha_2$, $\alpha_3$, and $\alpha_4$. The four slanting lines represent

```
VI(V, ε):
1.   δ ← ε(1 − λ)/2λ.
2.   do {
3.        U ← DP-UPDATE(V).
4.        r ← max_b |U(b) − V(b)|;
5.        if (r > δ) V ← U.
6.   } while ( r > δ).
7.   return U.
```

Figure 2: Value Iteration for POMDPs.

the linear functions $\alpha_i \cdot b$ ($i=1, 2, 3, 4$) of $b$. The value function induced by the four vectors is represented by the three bold line segments at the top. Vector $\alpha_3$ is extraneous as its removal does not affect the induced function. All the other vectors are useful. The first segment of the line at the bottom is the witness region of $\alpha_1$, the second segment is that of $\alpha_2$, and the last segment is that of $\alpha_4$.

### 2.5 Finite Representation of Value Functions and Value Iteration

A value function $V$ is *represented* by a set of vectors if it equals the value function induced by the set. When a value function is representable by a finite set of vectors, there is a unique parsimonious set of vectors that represents the function (Littman *et al.* 1995a).

Sondik (1971) has shown that if a value function $V$ is representable by a finite set of vectors, then so is the value function $TV$. The process of obtaining the parsimonious representation for $TV$ from the parsimonious representation of $V$ is usually referred to as *dynamic programming (DP) update*. Let $\mathcal{V}$ be the parsimonious set of vectors that represents $V$. For convenience, we use $T\mathcal{V}$ to denote the parsimonious set of vectors that represents $TV$.

In practice, value iteration for POMDPs is not carried out directly in terms of value functions themselves. Rather, it is carried out in terms of sets of vectors that represent the value functions (Figure 2). One begins with an initial set of vectors $\mathcal{V}$. At each iteration, one performs a DP update on the previous parsimonious set $\mathcal{V}$ of vectors and obtains a new parsimonious set of vectors $\mathcal{U}$. One continues until the Bellman residual $\max_b |\mathcal{U}(b) − \mathcal{V}(b)|$, which is determined by solving a sequence of linear programs, falls below a threshold.

## 3. Point-Based DP Update: The Idea

This section explains the intuitions behind point-based DP update. We begin with the so-called backup operator.

### 3.1 The Backup Operator

Let $\mathcal{V}$ be a set of vectors and $b$ be a belief state. The *backup operator* constructs a new vector in three steps:

1. For each action $a$ and each observation $z$, find the vector in $\mathcal{V}$ that has maximum inner product with $b_a^z$ — the belief state for the case when $z$ is observed after executing action $a$ in belief state $b$. If there are more than one such vector, break ties lexicographically (Littman 1996). Denote the vector found by $\beta_{a,z}$.

2. For each action $a$, construct a vector $\beta_a$ by:

$$\beta_a(s) = r(s, a) + \gamma \sum_{z,s'} P(s', z|s, a)\beta_{a,z}(s'), \forall s \in \mathcal{S}.$$

3. Find the vector, among the $\beta_a$'s, that has maximum inner product with $b$. If there are more than one such vector, break ties lexicographically. Denote the vector found by $\texttt{backup}(b, \mathcal{V})$.

It has been shown (Smallwood and Sondik 1973, Littman 1996) that $\texttt{backup}(b, \mathcal{V})$ is a member of $T\mathcal{V}$ — the set of vectors obtained by performing DP update on $\mathcal{V}$. Moreover, $b$ is a witness point for $\texttt{backup}(b, \mathcal{V})$.

The above fact is the corner stone of several DP update algorithms. The one-pass algorithm (Sondik 1971), the linear-support algorithm (Cheng 1988), and the relaxed-region algorithm (Cheng 1988) operate in the following way: They first systematically search for witness points of vectors in $T\mathcal{V}$ and then obtain the vectors using the backup operator. The witness algorithm (Kaelbling *et al.* 1998) employs a similar idea.

## 3.2 Point-Based DP Update

Systematically searching for witness points for all vectors in $T\mathcal{V}$ is computationally expensive. Point-based DP update does not do this. Instead, it uses heuristics to come up with a collection of belief points and backs up on those points. It might miss witness points for some of the vectors in $T\mathcal{V}$ and hence is an approximation of standard DP update.

Obviously, backing up on different belief states might result in the same vector. In other words, $\texttt{backup}(b, \mathcal{V})$ and $\texttt{backup}(b', \mathcal{V})$ might be equal for two different belief states $b$ and $b'$. As such, it is possible that one gets only a few vectors after many backups. One issue in the design of point-based DP update is to avoid this. We address this issue using witness points.

Point-based DP update assumes that one knows a witness point for each vector in its input set. It backs up on those points.[1] The rationale is that witness points for vectors in a given set "scatter all over the belief space" and hence the chance of creating duplicate vectors is low. Our experiments have confirmed this intuition.

The assumption made by point-based DP update is reasonable because its input is either the output of a standard DP update or another point-based DP update. Standard DP update computes, as by-products, a witness point for each of its output vectors. As will be seen later, point-based DP update also shares this property by design.

## 3.3 The Use of Point-Based DP Update

As indicated in the introduction, we propose to perform point-based DP update a number of times in between two standard DP updates. To be more specific, we propose to modify

---

1. As will be seen later, point-based DP update also backs up on other points.

```
VI1(V, ε):
1.   δ ← ε(1 − λ)/2λ.
2.   do {
3.       U ← DP-UPDATE(V).
4.       r ← max_b |U(b) − V(b)|;
5.       if (r > δ) V ← POINT-BASED-VI(U, δ).
6.   } while ( r > δ).
7.   return U.

POINT-BASED-VI(U, δ):
1.   do {
2.       V ← U.
3.       U ← POINT-BASED-DPU(V)
4.   } while (STOP(U, V, δ)= false).
5.   return V.
```

Figure 3: Modified Value Iteration for POMDPs.

value iteration in the way as shown in Figure 3. Note that the only change is at line 5. Instead of assigning $\mathcal{U}$ directly to $\mathcal{V}$, we pass it to a subroutine POINT-BASED-VI and assign the output of the subroutine to $\mathcal{V}$. The subroutine functions in the same way as value iteration, except that it performs point-based DP updates rather than standard DP updates. Hence we call it *point-based value iteration*.

Figure 4 illustrates the basic idea behind modified value iteration in contrast to value iteration. When the initial value function is properly selected,[2] the sequence of value functions produced by value iteration converges monotonically to the optimal value function. Convergence usually takes a long time partially because standard DP updates, indicated by fat upward arrows, are computationally expensive. Modified value iteration interleaves standard DP updates with point-based DP updates, which are indicated by the thin upward arrows. Point-based DP update does not improve a value function as much as standard DP update. However, its complexity is much lower. As a consequence, modified value iteration can hopefully converge in less time.

The idea of interleaving standard DP updates with approximate updates that back up at a finite number of belief points is due to Cheng (1988). Our work differs from Cheng's method mainly in the way we select the belief points. A detailed discussion of the differences will be given in Section 8.

The modified value iteration algorithm raises three issues. First, what stopping criterion do we use for point-based value iteration? Second, how can we guarantee the stopping criterion can eventually be satisfied? Third, how do we guarantee the convergence of the modified value iteration algorithm itself? To address those issues, we introduce the concept of uniformly improvable value functions.

---

2. We will show how in Section 5.5.

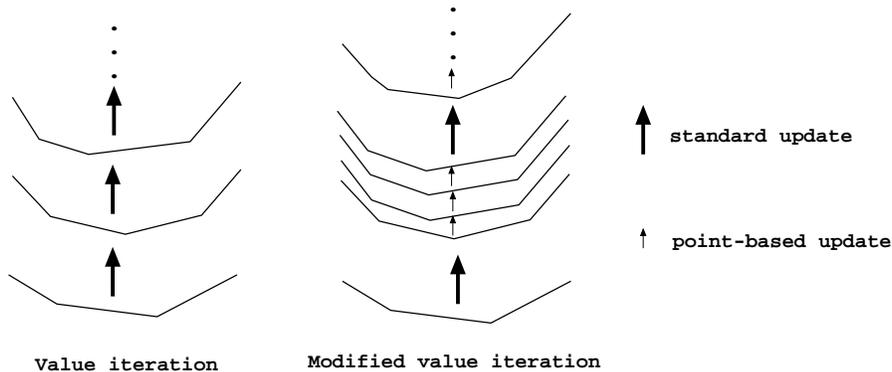

Figure 4: Illustration of the Basic Idea behind Modified Value Iteration.

## 4. Uniformly Improvable Value Functions

Suppose $V$ and $U$ are two value functions. We say that $U$ *dominates* $V$ and write $V \leq U$ if $V(b) \leq U(b)$ for every belief state $b$. A value function $V$ is said to be *uniformly improvable* if $V \leq TV$. A set $\mathcal{U}$ of vectors *dominates* another set $\mathcal{V}$ of vectors if the value function induced by $\mathcal{U}$ dominates that induced by $\mathcal{V}$. A set of vectors is *unformly improvable* if the value function it induces is.

**Lemma 1** *The operator $T$ is isotone in the sense that for any two value functions $V$ and $U$, $V \leq U$ implies $TV \leq TU$.* □

This lemma is obvious and is well known in the MDP community (Puterman 1990). Nonetheless, it enables us to explain the intuition behind the term "uniformly improvable". Suppose $V$ is a uniformly improvable value function and suppose value iteration starts with $V$. Then the sequence of value functions generated is monotonically increasing and converges to the optimal value function $V^*$. This implies $V \leq TV \leq V^*$. That is, $TV(b)$ is closer to $V^*(b)$ than $V(b)$ for all belief states $b$.

The following lemma will be used later to address the issues listed at the end of the previous section.

**Lemma 2** *Consider two value functions $V$ and $U$. If $V$ is uniformly improvable and $V \leq U \leq TV$, then $U$ is also uniformly improvable.*

**Proof**: Since $V \leq U$, we have $TV \leq TU$ by Lemma 1. We also have the condition $U \leq TV$. Consequently, $U \leq TU$. That is, $U$ is uniformly improvable. □

**Corollary 1** *If value function $V$ is uniformly improvable, so is $TV$.* □

## 5. Point-Based DP Update: The Algorithm

Point-based DP update is an approximation of standard DP update. When designing point-based DP update, we try to strike a balance between quality of approximation and computational complexity. We also need to guarantee that the modified value iteration algorithm converges.

## 5.1 Backing Up on Witness Points of Input Vectors

Let $\mathcal{V}$ be a set of vectors on which we are going to perform point-based DP update. As mentioned earlier, we can assume that we know a witness point for each vector in $\mathcal{V}$. Denote the witness point for a vector $\alpha$ by $\mathtt{w}(\alpha)$. Point-based DP update first backs up on these points and thereby obtains a new set of vectors. To be more specific, it begins with the following subroutine:

> backUpWitnessPoints($\mathcal{V}$):
> 1. $\mathcal{U} \leftarrow \emptyset$.
> 2. **for** each $\beta \in \mathcal{V}$
> 3.     $\alpha \leftarrow \mathtt{backup}(\mathtt{w}(\beta), \mathcal{V})$.
> 4.     **if** $\alpha \notin \mathcal{U}$
> 5.         $\mathtt{w}(\alpha) \leftarrow \mathtt{w}(\beta)$.
> 6.         $\mathcal{U} \leftarrow \mathcal{U} \cup \{\alpha\}$.
> 7. return $\mathcal{U}$.

In this subroutine, line 4 makes sure that the resulting set $\mathcal{U}$ contains no duplicates and line 5 takes note of the fact that $\mathtt{w}(\beta)$ is also a witness point for $\alpha$ (w.r.t $T\mathcal{V}$).

## 5.2 Retaining Uniform Improvability

To address convergence issues, we assume that the input to point-based DP update is uniformly improvable and require its output to be also uniformly improvable. We will explain later how the assumption can be facilitated and how the requirement guarantees convergence of the modified value iteration algorithm. In this subsection, we discuss how the requirement can be fulfilled.

Point-based DP update constructs new vectors by backing up on belief points and the new vectors are all members of $T\mathcal{V}$. Hence the output of point-based DP update is trivially dominated by $T\mathcal{V}$. If the output also dominates $\mathcal{V}$, then it must be uniformly improvable by Lemma 2. The question is how to guarantee that the output dominates $\mathcal{V}$.

Consider the set $\mathcal{U}$ resulted from $\mathtt{backUpWitnessPoints}$. If it does not dominate $\mathcal{V}$, then there must exist a belief state $b$ such $\mathcal{U}(b) < \mathcal{V}(b)$. Consequently, there must exist a vector $\beta$ in $\mathcal{V}$ such that $\mathcal{U}(b) < \beta \cdot b$. This gives us the following subroutine for testing whether $\mathcal{U}$ dominates $\mathcal{V}$ and for, when this is not the case, adding vectors to $\mathcal{U}$ so that it does. The subroutine is called $\mathtt{backUpLPPoints}$ because belief points are found by solving linear programs.

> backUpLPPoints($\mathcal{U}, \mathcal{V}$):
> 1. **for** each $\beta \in \mathcal{V}$
> 2.     **do** {
> 3.         $b \leftarrow \mathtt{dominanceCheck}(\beta, \mathcal{U})$.
> 4.         **if** $b \neq \mathtt{NULL}$,
> 5.             $\alpha \leftarrow \mathtt{backup}(b, \mathcal{V})$.

| | |
|---|---|
| 6. | $\texttt{w}(\alpha) \leftarrow b$. |
| 7. | $\mathcal{U} \leftarrow \mathcal{U} \cup \{\alpha\}$. |
| 8. | $\}$ **while** $(b \neq \texttt{NULL})$. |

The subroutine examines vectors in $\mathcal{V}$ one by one. For each $\beta$ in $\mathcal{V}$, it calls another subroutine `dominanceCheck` to try to find a belief point $b$ such that $\mathcal{U}(b) < \beta \cdot b$. If such a point is found, it backs up on it, resulting in a new vector $\alpha$ (line 5). By the property of the backup operator, $b$ is a witness point of $\alpha$ w.r.t $T\mathcal{V}$ (line 6). There cannot be any vector in $\mathcal{U}$ that equals $\alpha$.[3] Consequently, the vector is simply added to $\mathcal{U}$ without checking for duplicates (line 7). The process repeats for $\beta$ until `dominanceCheck` returns `NULL`, that is when there are no belief points $b$ such that $\mathcal{U}(b) < \beta \cdot b$. When `backUpLPPoints` terminates, we have $\mathcal{U}(b) \geq \beta \cdot b$ for any vector $\beta$ in $\mathcal{V}$ and any belief point $b$. Hence $\mathcal{U}$ dominates $\mathcal{V}$.

The subroutine `dominanceCheck`$(\beta, \mathcal{U})$ first checks whether there exists a vector $\alpha$ in $\mathcal{U}$ that *pointwise dominates* $\beta$, that is $\alpha(s) \geq \beta(s)$ for all states $s$. If such an $\alpha$ exists, it returns `NULL` right away. Otherwise, it solves the following linear program LP$(\beta, \mathcal{U})$. It returns the solution point $b$ when the optimal value of the objective function is positive and returns `NULL` otherwise:[4]

> LP$(\beta, \mathcal{U})$:
> 1. Variables: $x$, $b(s)$ for each state $s$
> 2. Maximize: $x$.
> 3. Constraints:
> 4. $\quad \sum_s \beta(s)b(s) \geq x + \sum_s \alpha(s) \cdot b(b)$ for all $\alpha \in \mathcal{U}$
> 5. $\quad \sum_s b(s) = 1$, $b(s) \geq 0$ for all states $s$.

## 5.3 The Algorithm

Here is the complete description of point-based DP update. It first backs up on the witness points of the input vectors. Then, it solves linear programs to identify more belief points and backs up on them so that its output dominates its input and hence is uniformly improvable.

> POINT-BASED-DPU$(\mathcal{V})$:
> 1. $\mathcal{U} \leftarrow$ `backUpWitnessPoints`$(\mathcal{V})$
> 2. `backUpLPPoints`$(\mathcal{U}, \mathcal{V})$
> 3. return $\mathcal{U}$.

In terms of computational complexity, point-based DP update performs exactly $|\mathcal{V}|$ backups in the first step and no more than $|T\mathcal{V}|$ backups in the second step. It solves linear programs only in the second step. The number of linear programs solved is upper bounded by $|T\mathcal{V}| + |\mathcal{V}|$ and is usually much smaller than the bound. The numbers of constraints in the linear programs are upper bounded by $|T\mathcal{V}| + 1$.

---

3. Since $b$ is a witness of $\alpha$ w.r.t $T\mathcal{V}$, we have $\alpha \cdot b = T\mathcal{V}(b)$. Since $\mathcal{V}$ is uniformly improvable, we also have $T\mathcal{V}(b) \geq \mathcal{V}(b)$. Together with the obvious fact that $\mathcal{V}(b) \geq \beta \cdot b$ and the condition $\beta \cdot b > \mathcal{U}(b)$, we have $\alpha \cdot b > \mathcal{U}(b)$. Consequently, there cannot be any vector in $\mathcal{U}$ that equals $\alpha$.

4. In our actual implementation, the solution point $b$ is used for backup even when the optimal value of the objective function is negative. In this case, duplication check is needed.

There are several algorithms for standard DP update. Among them, the incremental pruning algorithm (Zhang and Liu 1997) has been shown to be the most efficient both theoretically and empirically (Cassandra *et al.* 1997). Empirical results (Section 6) reveal that point-based DP update is much less expensive than incremental pruning on a number of test problems. It should be noted, however, that we have not proved this is always the case.

### 5.4 Stopping Point-Based Value Iteration

Consider the do-while loop of `POINT-BASED-VI` (Figure 2). Starting from an initial set of vectors, it generates a sequence of sets. If the initial set is uniformly improvable, then the value functions represented by the sets are monotonically increasing and are upper bounded by the optimal value function. As such, they converge to a value function (which is not necessarily the optimal value function). The question is when to stop the do-while loop.

A straightforward method would be to compute the distance $\max_b |\mathcal{U}(b) - \mathcal{V}(b)|$ between two consecutive sets $\mathcal{U}$ and $\mathcal{V}$ and stop when the distance falls below a threshold. To compute the distance, one needs to solve $|\mathcal{U}| + |\mathcal{V}|$ linear programs, which is time consuming. We use a metric that is less expensive to compute. To be more specific, we stop the do-while loop when

$$\max_{\alpha \in \mathcal{U}} |\mathcal{U}(\mathbf{w}(\alpha)) - \mathcal{V}(\mathbf{w}(\alpha))| \leq \delta_1 \delta.$$

In words, we calculate the maximum difference between $\mathcal{U}$ and $\mathcal{V}$ at the witness points of vectors in $\mathcal{U}$ and stop the do-while loop when this quantity is no larger than $\delta_1 \delta$. Here $\delta$ is the threshold on the Bellman residual for terminating value iteration and $\delta_1$ is a number between 0 and 1. In our experiments, we set it at 0.1.

### 5.5 Convergence of Modified Value Iteration

Let $\mathcal{V}_n$ and $\mathcal{V}'_n$ be sets of vectors respectively generated by VI (Figure 1) and VI1 (Figure 2) at line 3 in iteration $n$. Suppose the initial set is uniformly improvable. Using Lemma 2 and Corollary 1, one can prove by induction that $\mathcal{V}_n$ and $\mathcal{V}'_n$ are uniformly improvable for all $n$ and their induced value functions increase with $n$. Moreover, $\mathcal{V}'_n$ dominates $\mathcal{V}_n$ and is dominated by the optimal value function. It is well known that $\mathcal{V}_n$ converges to the optimal value function. Therefore, $\mathcal{V}'_n$ must also converge to the optimal value function.

The question now is how to make sure that the initial set is uniformly improvable. The following lemma answers this question.

**Lemma 3** *Let $m = \min_{s,a} r(s, a)$, $c = m/(1 - \lambda)$, and $\alpha_c$ be the vector whose components are all $c$. Then the singleton set $\{\alpha_c\}$ is uniformly improvable.*

**Proof**: Use $V$ to denote the value function induced by the singleton set. For any belief state $b$, we have

$$TV(b) \quad = \quad \max_a [r(b, a) + \lambda \sum_z P(z|b, a) V(b_z^a)]$$

$$
\begin{aligned}
&= \max_a [r(b,a) + \lambda \sum_z P(z|b,a)c] \\
&= \max_a [r(b,a) + \lambda m/(1-\lambda)] \\
&\geq m + \lambda m/(1-\lambda) \\
&= m/(1-\lambda) = V(b).
\end{aligned}
$$

Therefore the value function, and hence the singleton set, is uniformly improvable. □

Experiments (Section 6) have shown that VI1 is more efficient VI on a number of test problems. It should be noted, however, that we have not proved this is always the case. Moreover, complexity results by Papadimitriou and Tsitsiklis (1987) implies that the task of finding $\epsilon$-optimal policies for POMDPs is PSPACE-complete. Hence, the worst-case complexity should remain the same.

### 5.6 Computing the Bellman Residual

In the modified value iteration algorithm, the input $\mathcal{V}$ to standard DP update is always uniformly improvable. As such, its output $\mathcal{U}$ dominates its input. This fact can be used to simplify the computation of the Bellman residual. As a matter of fact, the Bellman residual $\max_b |\mathcal{U}(b) - \mathcal{V}(b)|$ reduces $\max_b (\mathcal{U}(b) - \mathcal{V}(b))$.

To compute the latter quantity, one goes through the vectors in $\mathcal{U}$ one by one. For each vector, one solves the linear program $\mathsf{LP}(\alpha, \mathcal{V})$. The quantity is simply the maximum of the optimal values of the objective functions of the linear programs. Without uniformly improvability, we would have to repeat the process one more time with the roles of $\mathcal{V}$ and $\mathcal{U}$ exchanged.

## 6. Empirical Results and Discussions

Experiments have been conducted to empirically determine the effectiveness of point-based DP update in speeding up value iteration. Eight problems are used in the experiments. In the literature, the problems are commonly referred to as 4x3CO, Cheese, 4x4, Part Painting, Tiger, Shuttle, Network, and Aircraft ID. We obtained the problem files from Tony Cassandra. Information about their sizes is summarized in the following table.

| Problem | $|\mathcal{S}|$ | $|\mathcal{Z}|$ | $|\mathcal{A}|$ | Problem | $|\mathcal{S}|$ | $|\mathcal{Z}|$ | $|\mathcal{A}|$ |
|---------|------|------|------|---------|------|------|------|
| 4x3CO | 11 | 4 | 11 | Cheese | 11 | 4 | 7 |
| 4x4 | 16 | 2 | 4 | Painting | 4 | 4 | 2 |
| Tiger | 2 | 2 | 3 | Shuttle | 8 | 2 | 3 |
| Network | 7 | 2 | 4 | Aircraft ID | 12 | 5 | 6 |

The effectiveness of point-based DP update is determined by comparing the standard value iteration algorithm VI and the modified value iteration algorithm VI1. The implementation of standard value iteration used in our experiments is borrowed from Hansen. Modified value iteration is implemented on top of Hansen's code.[5] The discount factor is set at 0.95 and round-off precision is set at $10^{-6}$. All experiments are conducted on an UltraSparc II machine.

---

5. The implementation is available on request.

Table 1 shows the amounts of time VI and VI1 took to compute 0.01-optimal policies for the test problems. We see that VI1 is consistently more efficient than VI, especially on the larger problems. It is about 1.3, 2.8, 5, 62, 141, 173, and 49 times faster than VI on the first seven problems respectively. For the Aircraft ID problem, VI1 was able to compute a 0.01-optimal policy in less than 8 hours, while VI was only able to produce a 33-optimal policy after 40 hours.

|      | 4x3CO | Cheese | 4x4   | Paint | Tiger | Shuttle | Network | Aircraft |
|------|-------|--------|-------|-------|-------|---------|---------|----------|
| VI   | 3.2   | 13.9   | 27.15 | 37.84 | 79.14 | 5,199   | 12,478  | -        |
| VI1  | 2.4   | 5.0    | 5.30  | .61   | .56   | 30      | 253     | 27,676   |

Table 1: Time for Computing 0.01-Optimal Policies in Seconds.

Various other statistics are given in Table 2 to highlight computational properties of VI1 and to explain its superior performance. The numbers of standard DP updates carried out by VI and VI1 are shown at rows 1 and 3. We see that VI1 performed no more than 5 standard updates on the test problems, while VI performed more than 125. This indicates that point-based update is very effective in cutting down the number of standard updates required to reach convergence. As a consequence, VI1 spent much less time than VI in standard updates (row 2 and 4).[6]

|         | Problem  | 4x3CO | Cheese | 4x4   | Paint | Tiger | Shuttle | Network |
|---------|----------|-------|--------|-------|-------|-------|---------|---------|
| VI      | DPU #    | 125   | 129    | 130   | 127   | 163   | 174     | 214     |
|         | Time     | 2.00  | 7.63   | 17.83 | 33.39 | 70.44 | 3,198   | 8,738   |
| VI1     | DPU #    | 4     | 4      | 3     | 3     | 3     | 5       | 5       |
|         | Time     | .05   | .09    | .15   | .21   | .09   | 13      | 82      |
|         | PBDPU #  | 377   | 219    | 173   | 244   | 515   | 455     | 670     |
|         | Time     | 2.32  | 4.86   | 5.09  | .37   | .45   | 10      | 139     |
| Quality Ratio   |  | .33   | .58    | .74   | .51   | .31   | 0.31    | .32     |
| Complexity Ratio |  | .38   | .37    | .21   | .0057 | .002  | .0012   | .005    |

Table 2: Detailed Statistics.

Row 5 shows the numbers of point-based updates carried out by VI1. We see that those numbers are actually larger than the numbers of standard updates performed by VI. This is expected. To see why, recall that point-based update is an approximation of standard update. Let $\mathcal{V}$ be a set of vectors that is uniformly improvable. Use $T'\mathcal{V}$ to denote the sets of vectors resulted from performing point-based update on $\mathcal{V}$. For any belief state $b$, we have $\mathcal{V}(b) \leq T'\mathcal{V}(b) \leq T\mathcal{V}(b)$. This means that point-based update improves $\mathcal{V}$ but not as much as standard update. Consequently, the use of point-based update increases the *total*

---

6. Note that times shown there do not include time for testing the stopping condition.

*number of iterations*, i.e the number of standard updates plus the number of point-based updates.

Intuitively, the better point-based update is as an approximation of standard update, the less the difference between the total number iterations that VI1 and VI need take. So, the ratio between those two numbers in a problem can be used, to certain extent, as a measurement of the quality of point-based update in that problem. We shall refer to it as the *quality ratio* of point-based update. Row 7 shows the quality ratios in the seven test problems. We see that the quality of point-based update is fairly good and stable across all the problems.

Row 8 shows, for each test problem, the ratio between the average time of a standard update performed by VI and that of a point-based update performed by VI1. Those ratios measure, to certain extent, the complexity of point-based update relative to standard update and hence will be referred to as the *complexity ratios* of point-based update. We see that, as predicted by the analysis in Section 5.3, point-based update is consistently less expensive than standard update. The differences are more than 200 times in the last four problems.

In summary, the statistics suggest that the quality of point-based update relative to standard update is fairly good and stable and its complexity is much lower. Together with the fact that point-based update can drastically reduces the number of standard updates, those explain the superior performance of VI1.

To close this section, let us note that while VI finds policies with quality[7] very close to the predetermined criterion, VI1 usually finds much better ones (Table 3). This is because VI checks policy quality after each (standard) update, while VI1 does not do this after point-based updates.

| Problem | 4x3CO | Cheese | 4x4 | Paint | Tiger | Shuttle | Network |
|---------|-------|--------|------|-------|-------|---------|---------|
| VI | .0095 | .0099 | .0099 | .01 | .0098 | .0097 | .0098 |
| VI1 | .0008 | .0008 | .0009 | .0007 | .0007 | .00015 | .001 |

Table 3: Quality of Policies Found by VI and VI1.

## 7. Variations of Point-Based DP Update

We have studied several possible variations of point-based update. Most of them are based on ideas drawn from the existing literature. None of the variations were able to significantly enhance the effectiveness of the algorithm in accelerating value iteration. Nonetheless a brief discussion of some of them is still worthwhile. The discussion provides further insights about the algorithm and shows how it compares to some of the related work to be discussed in detail in the next section.

The variations can be divide into two categories: those aimed at improving the quality of point-based update and those aimed at reducing complexity. We shall discuss them one by one.

---

7. Quality of a policy is estimated using the Bellman residual.

## 7.1 Improving the Quality of Point-Based DP Update

A natural way to improve the quality of point-based update is to back up on additional belief points. We have explored the use of randomly generated points (Cassandra 1998a), additional by-product points, and projected points (Hauskrecht 2000). Here additional by-product points refer to points generated at various stages of standard update, excluding the witness points that are already being used. Projected points are points that are reachable in one step from points that have given rise to useful vectors.

Table 4 shows, for each test problem, the number of standard updates and the amount of time that VI1 took with and without using projected points. We see that the use of projected points did reduce the number of standard updates by one in 4x3CO, Cheese, and Shuttle. However, it increased the time complexity in all test problems except for Network. The other two kinds of points and combinations of the three did not significantly improve VI1 either. On the contrary, they often significantly degraded the performance of VI1.

|       | 4x3CO | Cheese | 4x4 | Paint | Tiger | Shuttle | Network | Aircraft |
|-------|-------|--------|-----|-------|-------|---------|---------|----------|
| w/o   | 4     | 4      | 3   | 3     | 3     | 5       | 5       | 7        |
| with  | 3     | 3      | 3   | 3     | 3     | 4       | 5       | 7        |
| w/o   | 2.4   | 5.0    | 5.3 | .61   | .56   | 30      | 253     | 27,676   |
| with  | 3.2   | 5.6    | 7.4 | .69   | 2.3   | 33      | 140     | 35,791   |

Table 4: Number of Standard DP Updates and Time That VI1 Took With and Without Using Projected Points.

A close examination of experimental data reveals a plausible explanation. Point-based update, as it stands, can already reduce the number of standard updates down to a just few and among them the last two or three are the most time-consuming. As such, the possibility of further reducing the number standard updates is low and even when it is reduced, the effect is roughly to shift the most time-consuming standard updates earlier. Consequently, it is unlikely to achieve substantial gains. On the other hand, the use of additional points always increases overheads.

## 7.2 Reducing the Complexity of Point-Based DP Update

Solving linear programs is the most expensive operation in point-based update. An obvious way to speed up is to avoid linear programs. Point-based update solves linear programs and backs up on the belief points found so as to guarantee uniform improvability. If the linear programs are to be skipped, there must be some other way to guarantee uniform improvability. There is an easy solution to this problem. Suppose $\mathcal{V}$ is the set of vectors that we try to update and it is uniformly improvable. Let $\mathcal{U}$ be the set obtained from $\mathcal{V}$ by backing up only on the witness points, which can be done without solving linear programs. The set $\mathcal{U}$ might or might not be uniformly improvable. However, the union $\mathcal{V} \cup \mathcal{U}$ is guaranteed to be uniformly improvable. Therefore we can reprogram point-based update to return the union in hope to reduce complexity. The resulting variation will be called *non-LP point-based DP update*.

Another way to reduce complexity is to simplify the backup operator (Section 3.1) using the idea behind modified policy iteration (e.g., Puterman 1990). When backing up from a set of vectors $\mathcal{V}$ at a belief point, the operator considers all possible actions and picks the one that is optimal according to the $\mathcal{V}$-improving policy. To speed up, one can simply use the action found for the belief point by the previous standard update. The resulting operator will be called the *MPI backup operator*, where MPI stands for modified policy iteration. If $\mathcal{V}$ is the output of the previous standard update, the two actions often are the same. However, they are usually different if $\mathcal{V}$ is the result of several point-based updates following the standard update.

Table 5 shows, for each test problem, the number of standard updates and the amount of time that VI1 took when non-LP point-based update was used (together with the standard backup operator). Comparing the statistics with those for point-based update (Tables 1 and 2), we see that the number of standard updates is increased on all test problems and the amount of time is also increased except for the first three problems. Here are the plausible reasons. First, it is clear that non-LP point-based update does not improve a set of vectors as much as point-based update. Consequently, it is less effective in reducing the number of standard updates. Second, although it does not solve linear programs, non-LP point-based update produces extraneous vectors. This means that it might need to deal with a large number of vectors at later iterations and hence might not be as efficient as point-based update after all.

| 4x3CO | Cheese | 4x4 | Paint | Tiger | Shuttle | Network | Aircraft |
|---|---|---|---|---|---|---|---|
| 4 | 5 | 8 | 4 | 4 | 7 | 10 | 8 |
| 2.38 | 2.38 | 3.4 | .75 | .88 | 44 | 599 | 32,281 |

Table 5: Number of Standard DP Updates and Time That VI1 Took When Non-LP Point-Based Update is Used.

Extraneous vectors can be pruned. As a matter of fact, we did prune vectors that are pointwise-dominated by others (hence extraneous) in our experiments. This is inexpensive. Pruning of other extraneous vectors, however, requires the solution of linear programs and is expensive. In Zhang *et al.* (1999), we have discussed how this can be done the most efficient way. Still the results were not as good as those in Table 5. In that paper, we have also explored the combination of non-LP point-based update with the MPI backup operator. Once again, the results were not as good as those in Table 5. The reason is that the MPI backup operator further compromises the quality of point-based update.

The quality of non-LP point-based update can be improved by using the Gauss-Seidel asynchronous update (Denardo 1982). Suppose we are updating a set $\mathcal{V}$. The idea is to, after a vector is created by backup, add a copy of the vector to the set $\mathcal{V}$ right away. The hope is to increase the components of later vectors. We have tested this idea when preparing Zhang *et al.* (1999) and found that the costs almost always exceed the benefits. A reason is that asynchronous update introduces many more extraneous vectors than synchronous update.

In conclusion, point-based is conceptually simple and clean. When compared to its more complex variations, it seems to be the most effective in accelerating value iteration.

## 8. Related Work

Work presented in this paper has three levels: point-based DP update at the bottom, point-based value iteration in the middle, and modified value iteration at the top. In this section, we discuss previous relevant work at each of the three levels.

### 8.1 Point-Based DP Update and Standard DP Update

As mentioned in Section 3.1, point-based update is closely related to several exact algorithms for standard update, namely one-pass (Sondik 1971), linear-support (Cheng 1988), and relaxed-region (Cheng 1988). They all backup on a finite number of belief points. The difference is that these exact algorithms generate the points systematically, which is expensive, while point-based update generate the points heuristically.

There are several other exact algorithms for standard DP update. The enumeration/reduction algorithms (Monahan 1982, Eagle 1984) and incremental pruning (Zhang and Liu 1997, Cassandra *et al.* 1997) first generate a set of vectors that are not parsimonious and then prune extraneous vectors by solving linear programs. Point-based DP update never generates extraneous vectors. It might generate duplicate vectors. However, duplicates are pruned without solving linear programs. The witness algorithm (Kaelbling *et al.* 1998) has two stages. In the first stage, it considers actions one by one. For each action, it constructs a set of vectors based on a finite number of systematically generated belief points using an operator similar to the backup operator. In the second stage, vectors for different actions are pooled together and extraneous vectors are pruned.

There are proposals to carry out standard update approximately by dropping vectors that are marginally useful (e.g., Kaelbling *et al.* 1998, Hansen 1998). Here is one idea along this line that we have empirically evaluated. Recall that to achieve $\epsilon$-optimality, the stopping threshold for the Bellman residual should be $\delta = \epsilon(1 - \lambda)/2\lambda$. Our idea is to drop marginally useful vectors at various stages of standard update while keeping the overall error under $\delta/2$ and to stop when the Bellman residual falls below $\delta/2$. It is easy to see that $\epsilon$-optimality is still guaranteed this way. We have also tried to start with a large error tolerance in hope to prune more vectors and gradually decrease the tolerance level to $\delta/2$. Reasonable improvements have been observed especially when one does not need quality of policy to be high. However such approximate updates are much more expensive than point-based updates. In the context of the modified value iteration algorithm, they are more suitable alternatives to standard updates than point-based update.

### 8.2 Point-Based Value Iteration and Value Function Approximation

Point-based value iteration starts with a set of vectors and generates a sequence of vector sets by repeatedly applying point-based update. The last set can be used to approximate the optimal value function.

Various methods for approximating the optimal value function have been developed previously.[8] We will compare them against point-based value iteration along two dimensions: (1) Whether they map one set of vectors to another, that is whether the can be interleaved with standard updates, and (2) if they do, whether they can guarantee convergence when interleaved with standard updates.

Lovejoy (1993) proposes to approximate the optimal value function $V^*$ of a POMDP using the optimal value function of the underlying Markov decision process (MDP). The latter is a function over the state space. So $V^*$ is being approximated by one vector. Littman *et al.* (1995b) extend this idea and approximate $V^*$ using $|\mathcal{A}|$ vectors, each of which corresponds to a Q-function of the underlying MDP. A further extension is recently introduced by Zubek and Dietterich (2000). Their idea is to base the approximation not on the underlying MDP, rather on a so-called even-odd POMDP that is identical to the original POMDP except that the state is fully observable at even time steps. Platzman (1980) suggests approximating $V^*$ using the value functions of one or more fixed suboptimal policies that are constructed heuristically. Those methods do not start with a set of vectors and hence do not map a set of vectors to another. However, they can easily be adapted to do so. However, they all put a predetermined limit on the number of output vectors. Consequently, convergence is not guaranteed when they are interleaved with standard updates.

Fast informed bound (Hauskrecht 1997a), Q-function curve fitting (Littman *et al.* 1995b), and softmax curve fitting (Parr and Russell 1995) do map a set of vectors to another. However, they differ drastically from point-based value iteration and from each other in their ways of deriving the next set of vectors from the current one. Regardless of the size of the current set, fast informed bound and Q-function curve fitting always produces $|\mathcal{A}|$ vectors, one for each action. In softmax curve fitting, the number of vectors is also determined a priori, although it is not necessarily related to the number of actions. Those methods can be interleaved with standard DP updates. Unlike point-based value iteration, they themselves may not converge (Hauskrecht 2000). Even in cases where they do converge themselves, the algorithms resulting from interleaving them with standard updates do not necessarily converge due to the a priori limits on the number of vectors.

Grid-based interpolation/extrapolation methods (Lovejoy 1991, Brafman 1997, Hauskrecht 1997b) approximate value functions by discretizing the belief space using a fixed or variable grid and by maintaining values only for the grid points. Values at non-grid points are estimated by interpolation/extrapolation when needed. Such methods cannot be interleaved with standard DP updates because they do not work with sets of vectors.

There are grid-based methods that work with sets of vectors. Lovejoy's method to lower bound the optimal value function (Lovejoy 1991), for instance, falls into this category. This method is actually identical to point-based value iteration except for the way it derives the next set of vectors from the current one. Instead of using point-based update, it backs up on grid points in a regular grid. Convergence of this method is not guaranteed. The algorithm resulting from interleaving it with standard updates may not converge either.

---

8. Hauskrecht (2000) has conducted an extensive survey on previous value function approximation methods and has empirically compared them in terms of, among other criteria, complexity and quality. It would be interesting to also include point-based value iteration in the empirical comparison. This is not done in the present paper because our focus is on using point-based value iteration to speed value iteration, rather than using as a value function approximation method.

The incremental linear-function method (Hauskrecht 2000) roughly corresponds to a variation of point-based value iteration that uses non-LP point-based update (Section 7.2) augmented with the Gauss-Seidel asynchronous update. The method does not have access to witness point. So it starts, for the purpose of backup, with extreme points of the belief space and supplement them with projected points. This choice of points appears poor because it leads to a large number of vectors and consequently the backup process is "usually stopped well before" convergence (Hauskrecht 2000).

## 8.3 Previous Work Related to Modified Value Iteration

The basic idea of our modified value iteration algorithm VI1 is to add, in between two consecutive standard updates, operations that are inexpensive. The hope is that those operations can significantly improve the quality of a vector set and hence reduce the number of standard updates.

Several previous algorithms work in the same fashion. The differences lie in the operations that are inserted between standard updates. The reward revision algorithm (White *et al.* 1989) constructs, at each iteration, a second POMDP based on the current set of vectors. It runs value iteration on the second POMDP for a predetermined number of steps. The output is used to modify the current set of vectors and the resulting set of vectors is fed to the next standard update.

Why is reward revision expected to speed up value iteration? Let $V$ be the value function represented by the current set of vectors. The second POMDP is constructed in such way that it shares the same optimal value function as the original POMDP if $V$ is optimal. As such, one would expect the two POMDPs to have similar optimal value functions if $V$ is close to optimal. Consequently, running value iteration on the second POMDP should improve the current value function. And it is inexpensive to do so because the second POMDP is fully observable.

Reward revision is conceptually much more complex than VI1 and seems to be less efficient. According to White *et al.* (1989), reward revision can, on average, reduce the number of standard updates by 80% and computational time by 85%. From Tables 1 and 2, we see that the differences between VI1 and VI are much larger.

The iterative discretization procedure (IDP) proposed by Cheng (1988) is very similar to VI1. There are two main differences. While VI1 uses point-based update, IDP uses non-LP point-based update. While point-based update in VI1 backs up on witness points and belief points found by linear programs, non-LP point-based update in IDP backs up on extreme points of witness regions found as by-products by Cheng's linear-support or relaxed region algorithms.

Cheng has conducted extensive experiments to determine the effectiveness of IDP in accelerating value iteration. It was found that IDP can cut the number of standard updates by as much as 55% and the amount of time by as much as 80%. Those are much less significant than the reductions presented in Tables 1 and 2.

Hansen's policy iteration (PI) algorithm maintains a policy in the form of a finite-state controller. Each node in the controller represents a vector. At each iteration, a standard update is performed on the set of vectors represented in the current policy. The resulting set of vectors is used to improve the current policy[9] and the improved policy is evaluated by solving a system of linear equations. This gives rise to a third set of vectors, which is fed to the next standard update.

We compared the performance of Hansen's PI algorithm to VI1. Table 6 shows, for each test problem, the number of standard updates and the amount of time the algorithm took. Comparing with the statistics for VI1 (Table 4), we see that PI performed more standard updates than VI1. This indicates that policy improvement/evaluation is less effective than point-based value iteration in cutting down the number of standard updates. In terms of time, PI is more efficient than VI1 on the first three problems but significantly less efficient on all other problems.

| 4x3CO | Cheese | 4x4 | Paint | Tiger | Shuttle | Network | Aircraft |
|-------|--------|-----|-------|-------|---------|---------|----------|
| 3 | 7 | 7 | 10 | 14 | 9 | 18 | 9 |
| .14 | .87 | 3.4 | 3.8 | 4.5 | 60 | 1,109 | 66,964 |

Table 6: Number of Standard Updates and Time That PI Took to Compute 0.01-Optimal Policies.

It might be possible to combine VI1 and PI. To be more specific, one can probably insert a policy improvement/evaluation step between two point-based updates in point-based value iteration (Figure 2). This should accelerate point-based value iteration and hence VI1. This possibility and its benefits are yet to be investigated.

## 9. Conclusions and Future Directions

Value iteration is a popular algorithm for finding $\epsilon$-optimal policies for POMDPs. It typically performs a large number of DP updates before convergence and DP updates are notoriously expensive. In this paper, we have developed a technique called point-based DP update for reducing the number of standard DP updates. The technique is conceptually simple and clean. It can easily be incorporated into most existing POMDP value iteration algorithms. Empirical studies have shown that point-based DP update can drastically cut down the number of standard DP updates and hence significantly speeding up value iteration. Moreover, point-based DP update compares favorably with its more complex variations that we can think also. It also compares favorably with policy iteration.

The algorithm presented this paper still requires standard DP updates. This limits its capability of solving large POMDPs. One future direction is to investigate the properties of point-based value iteration as an approximation algorithm by itself. Another direction is to design efficient algorithms for standard DP updates in special models. We are currently exploring the latter direction.

---

9. In Hansen's writings, policy improvement includes DP update as a substep. Here DP update is not considered part of policy improvement.

## Acknowledgments

Research is supported by Hong Kong Research Grants Council Grant HKUST6125/98E. The authors thank Tony Cassandra and Eric Hansen for sharing with us their programs. We are also grateful for the three anonymous reviewers who provided insightful comments and suggestions on an earlier version of this paper.